# Improved Dynamic Schedules for Belief Propagation

**Charles Sutton and Andrew McCallum**
Department of Computer Science
University of Massachusetts
Amherst, MA 01003 USA
{casutton,mccallum}@cs.umass.edu

## Abstract

Belief propagation and its variants are popular methods for approximate inference, but their running time and even their convergence depend greatly on the schedule used to send the messages. Recently, dynamic update schedules have been shown to converge much faster on hard networks than static schedules, namely the residual BP schedule of Elidan et al. [2006]. But that RBP algorithm wastes message updates: many messages are computed solely to determine their priority, and are never actually performed. In this paper, we show that estimating the residual, rather than calculating it directly, leads to significant decreases in the number of messages required for convergence, and in the total running time. The residual is estimated using an upper bound based on recent work on message errors in BP. On both synthetic and real-world networks, this dramatically decreases the running time of BP, in some cases by a factor of five, without affecting the quality of the solution.

## 1 INTRODUCTION

Many popular approximate inference methods, such as belief propagation, its generalizations, including EP [Minka, 2001] and GBP [Yedidia et al., 2000], and structured mean-field methods [Jordan et al., 1999], consist of a set of equations which are iterated to find a fixed point. The fixed-point updates are not usually guaranteed to converge. The schedule for propagating the updates can make a crucial difference both to how long the updates take to converge, and even whether they converge at all.

Recently, *dynamic schedules*—in which the message values during inference are used to determine which update to perform next—have been shown to converge much faster on hard networks than static schedules [Elidan et al., 2006]. But the residual schedule proposed by Elidan et al., which we call *residual BP with lookahead one (RBP1L)*, determines a message's priority by actually computing it, which means that many message updates are "wasted", that is, they are computed solely for the purpose of computing their priority, and are never actually performed. A significant fraction of messages computed by RBP1L are wasted in this way.

In this paper, we show that rather than *computing* the residual of each pending message update, it is far more efficient to *approximate* it. Recent work [Ihler et al., 2004] has examined how a message error can be estimated as a function of its incoming errors. In our situation, the error arises because the incoming messages have been recomputed. The arguments from Ihler et al. apply also to the message residual, which leads to effective method for estimating the residual of a message, and to a dynamic schedule that is dramatically more efficient than RBP1L.

The contributions of this paper are as follows. First, we describe how the message residual can be upper-bounded by the residuals of its incoming messages (Section 3). We also describe a method for estimating the message residual when the factors themselves change (for example, from parameter updates), which leads to an intuitive method for initializing the residual estimates. Then we introduce a novel message schedule, which we call *residual BP with lookahead zero (RBP0L)* (Section 4). On several synthetic and real-world data sets, we show that RBP0L is as much as five times faster than RBP1L, while still finding the same solution (Section 5). Finally, we examine how to what extent the distance that a message changes in a single update predicts its distance to its final converged value (Section 5.3). We measure distance in several different ways, including the dynamic range of the error and the Bethe energy. Surprisingly, the difference in Bethe energy has almost no predictive value for whether a message update is nearing convergence.

## 2 BACKGROUND

Let $p(\mathbf{x})$ factorize according to an undirected factor graph $G$ [Kschischang et al., 2001] with factors $\{t_a(x_a)\}_{a=1}^{A}$, so that $p$ can be written as

$$p(\mathbf{x}) = \frac{1}{Z} \prod_a t_a(x_a), \qquad (1)$$

where $Z$ is the normalization constant

$$Z = \sum_{\mathbf{x}'} \prod_a t_a(x'_a). \qquad (2)$$

We will use the indices $a$ and $b$ to denote factors of $G$, and the indices $i$ and $j$ to denote variables. By $\{i \in a\}$ we mean the set of all variables $i$ in the domain of the factor $t_a$, and conversely by $\{b \ni i\}$, we mean the set of all factors $t_b$ that have variable $i$ in their domain.

Belief propagation (BP) is a popular approximate inference algorithm for factor graphs, dating to Pearl [1988]. The messages at iteration $k+1$ are computed by iterating the updates

$$\begin{aligned} m_{ai}^{(k+1)}(x_i) &\leftarrow \kappa \sum_{x_a \setminus x_i} t_a(x_a) \prod_{\{j \in a\} \setminus i} m_{ja}^{(k)}(x_j) \\ m_{ia}^{(k+1)}(x_i) &\leftarrow \kappa \prod_{\{b \ni i\} \setminus a} m_{bi}^{(k)}(x_i) \end{aligned} \qquad (3)$$

until a fixed point is reached. In the above, $\kappa$ is a normalization constant to ensure the message sums to 1. The initial messages $m^{(0)}$ are set to some arbitrary value, typically a uniform distribution.

We write the message updates in a generic fashion as

$$m_{cd}^{(k+1)}(x_{cd}) \leftarrow \kappa \sum_{x_c \setminus x_{cd}} t_a(x_c) \prod_{\{b \in N(c)\} \setminus d} m_{bc}^{(k)}(x_c), \qquad (4)$$

where $c$ and $d$ may be either factors or variables, as long as they are neighbors, $N(c)$ means the set of neighbors of $c$, and $t_a(x_c)$ is understood to be the identity if $c$ is a variable. This notation abstracts over whether a message is being sent from a factor or from a variable, which is convenient for describing message schedules.

In general, these updates may have multiple fixed points, and they are not guaranteed to converge. Convergent methods for optimizing the Bethe energy have been developed [Yuille and Rangarajan, 2001, Welling and Teh, 2001], but they are not used in practice both because they tend to be slower than iterating the messages (3), and because when the BP updates do not converge, it has been observed that the Bethe approximation is bad anyway.

Now we describe in more detail how the iterations are actually performed in a BP implementation. This level of detail will prove useful in the next section for understanding the behavior of dynamic BP schedules. A vector $\mathbf{m} = \{m_{cd}\}$ is maintained of all the messages, which is initialized to uniform. Then until the messages are converged, we iterate: A message $m_{cd}$ is selected according to the message update schedule. The new value $m'_{cd}$ is computed from its dependent messages in $\mathbf{m}$, according to (3). Finally, the old message $(c, d)$ in $\mathbf{m}$ is replaced with the newly computed value $m'_{cd}$.

The important part of this description is the distinction between when a message update is *computed* and when it is *performed*. When a message is computed, this means that its new value is calculated according to (3). When a message is performed, this means that the current message vector $\mathbf{m}$ is updated with the new value. Synchronous BP implementations compute all of the updates first, and then perform them all at once. Asynchronous BP implementations almost always perform an update as soon as it is computed, but it is possible to compute an update solely in order to determine its priority, and not perform the update until later. As we describe below, this is exactly the technique used by the Elidan et al. [2006] schedule.

## 3 ESTIMATING MESSAGE RESIDUALS

In this section, we describe how to compute an upper bound on the error of a message, which will be used as a priority for scheduling messages. We define the *error* $e_{cd}(x_{cd})$ of a message $m_{cd}(x_{cd})$ as its multiplicative distance from its previous value $m_{cd}^{(k)}(x_{cd})$, so that

$$m_{cd}(x_{cd}) = e_{cd}(x_{cd}) m_{cd}^{(k)}(x_{cd}). \qquad (5)$$

We define the *residual* of a message $m_{cd}(x_{cd})$ as the worst error over all assignments, that is,

$$r(m_{cd}) = \max_{x_{cd}} |\log e_{cd}(x_{cd})| = \max_{x_{cd}} \left| \log \frac{m_{cd}(x_{cd})}{m_{cd}^{(k)}(x_{cd})} \right|. \qquad (6)$$

This corresponds to using the infinity norm to measure the distance between log message vectors, that is, $\| \log m_{cd} - \log m_{cd}^{(k)} \|_\infty$.

An alternative error measure is the *dynamic range* of the error, which has been studied by Ihler et al. [2004]. This is

$$d(m_{cd}) = \max_{x_{cd}, x'_{cd}} \log \frac{e_{cd}(x_{cd})}{e_{cd}(x'_{cd})} \qquad (7)$$

Later we compare the residual and the dynamic error range as priority functions for message scheduling.

In the rest of this section, we show how to upper-bound the message errors in two different situations: when the values of a message's dependents change, and when the factors of the model change.

First, suppose that we have available a previously-computed message value for $m_{cd}^{(k)}(x_d)$, so that

$$m_{cd}^{(k)}(x_d) = \kappa \sum_{x_{cd}} t_c(x_c) \prod_{\{b \in N(c)\} \setminus d} m_{bc}^{(k)}(x_c), \quad (8)$$

and that now new messages $\{m_{bc}^{(k+1)}\}$ are available for the dependents. We wish to upper bound the residual $r_{cd}$ without actually repeating the update (8). Then the residual can be upper-bounded simply by the following:

$$r(m_{cd}) \leq \sum_{\{b \in N(c)\}} r(m_{bc}). \quad (9)$$

The full proof is given in the Appendix, but it is a straightforward application of the corresponding arguments from Ihler et al. for the dynamic range measure.

Now consider the second situation, when a factor $t_a$ changes. Define $e_a$ to be the multiplicative error in the factor, so that

$$t_a^{(k+1)}(x_a) = e_a(x_a) t_a^{(k)}(x_a). \quad (10)$$

Suppose we have already computed a message $m_{cd}^{(k)}$, so that in the current message vector

$$m_{cd}^{(k)}(x_d) = \sum_{x_{cd}} t_c^{(k)}(x_c) \prod_{\{b \in N(c)\} \setminus d} m_{bc}^{(k)}(x_c), \quad (11)$$

and as before we wish to upper bound $r(m_{cd})$. Then substitution into (6) yields

$$r(m_{cd}) \leq \max_{x_a} \frac{t_a^{(k+1)}(x_a)}{t_a^{(k)}(x_a)}. \quad (12)$$

## 4 DYNAMIC BP SCHEDULES

There has been little work in how to schedule the message updates (3). Recently, Elidan et al. [2006] showed that *dynamic schedules* are significantly superior to static schedules for BP, including several versions of the tree reparameterization schedule (TRP) [Wainwright et al., 2001].

### 4.1 RESIDUAL BP WITH LOOKAHEAD (RBP1L)

In this section, we describe the dynamic schedule proposed by Elidan et al. [2006]. They call their algorithm *residual belief propagation*, but in the next section we introduce a different BP schedule that also depends on the message residual. Therefore, to avoid confusion we refer to the Elidan et al. algorithm by the more specific name of *residual BP with lookahead one (RBP1L)*.

**Algorithm 1** RBP1L [Elidan et al., 2006]

**function** RBP1L ()
1: $\mathbf{m} \leftarrow$ uniform message array
2: $q \leftarrow$ INITIALPQ()
3: **repeat**
4:   $m_{bc} \leftarrow$ DEQUEUE($q$)
5:   $\mathbf{m}|_{bc} \leftarrow m_{bc}$ {Perform update.}
6:   **for all** $d$ in $\{d \in N(c)\} \setminus b$ **do**
7:     Compute update $m_{cd}$
8:     Remove any pending update $m_{cd}^{(k)}$ from $q$
9:     Add $m_{cd}$ to $q$ with priority $r(m_{cd})$
10:  **end for**
11: **until** messages converged

**function** INITIALPQ ()
1: $q \leftarrow$ empty priority queue
2: **for all** messages $(c, d)$ **do** {Initialize $q$}
3:   Compute update $m_{cd}$
4:   Add $m_{cd}$ to $q$ with priority $r(m_{cd})$
5: **end for**
6: **return** $q$

The basic idea in RBP1L (Algorithm 1) is that whenever a message $m_{cd}$ is pending for an update, the message is computed and placed on a priority queue to be performed. The priority of the message is the distance between its current value and its newly-computed value: the exact distance measure is not specified by Elidan et al., although they assume that it is based on a norm $\|m_{cd} - m_{cd}^{(k)}\|$ between the difference in message values. In this paper, we use the residual (6) between log message values.

The problem with this schedule can be seen in Lines 7–9 of Algorithm 1. When an update $m_{bc}$ is performed, each of its dependents $m_{cd}$ is recomputed and placed in the queue. If a previous update $m_{cd}^{(k)}$ was already pending in the queue, then that message is discarded. We refer to this as a "wasted" update. In Section 5, we see that this is a relatively common occurrence in RBP1L, so preventing this can yield to significant gains in convergence speed.

### 4.2 AVOIDING LOOKAHEAD (RBP0L)

In this section we present our dynamic schedule, *residual BP with lookahead zero (RBP0L)*. In Section 3 we showed that a residual can be upper-bounded by its sum of incoming residuals. The idea behind RBP0L is to use that upper bound as the message's priority, so that an update is never computed unless it will actually be performed. The full algorithm is given in Algorithm 2.

There are three fine points here. The first question is

**Algorithm 2** RBP0L

**function** RBP0L ()
1: **m** ← uniform message array
2: $T$ ← total residuals; initialized to 0
3: $q$ ← INITIALPQ()
4: **repeat**
5: 　$m_{bc}$ ← DEQUEUE($q$)
6: 　Compute update $m_{bc}$ and residual $r = r(m_{bc})$
7: 　$\mathbf{m}|_{bc}$ ← $m_{bc}$ {Perform update.}
8: 　For all $ab$, do $T(ab, bc)$ ← 0
9: 　For all $cd$, do $T(bc, cd)$ ← $T(bc, cd) + r$
10: 　**for all** $d$ in $\{d \in N(c)\} \setminus b$ **do**
11: 　　$v$ ← $\sum_a T(ac, cd)$
12: 　　Remove any pending update $(c, d)$ from $q$
13: 　　Add $m_{cd}$ to $q$ with priority $v$
14: 　**end for**
15: **until** messages converged

**function** INITIALPQ ()
1: $q$ ← empty priority queue
2: **for all** messages $(c, d)$ **do** {Initialize $q$}
3: 　Compute update $m_{cd}$
4: 　$v$ ← $\max_{x_c} |X_c| \left|\log t_c(x_c)\right|$
5: 　Add $m_{cd}$ to $q$ with priority $v$
6: **end for**
7: **return** $q$

how to update the residual estimate when a message $m_{bc}(x_c)$ is updated twice before one of its dependents $m_{cd}(x_d)$ is updated even once. In the most general case, each dependent may have actually seen a different version of $m_{bc}$ when it was last updated. Naively applying the bound (8) would suggest that we retain the version of $m_{bc}$ as it was when each of its dependents last saw it. But this becomes somewhat expensive in terms of memory. Instead, for each pair of messages $(b, c)$ and $(c, d)$ we maintain a total residual $T(bc, cd)$ of how much the message $m_{bc}$ has changed since $m_{cd}$ was last updated. Estimates of the priority of $m_{cd}$ are always computed using the total residual, rather than the single-update residual. (This preserves the upper-bound property of the residual estimates.)

The second question is how to initialize the residual estimates. Recall that the messages **m** are initialized to uniform. Imagine that those initial messages were obtained by starting with a factor graph in which all factors $t_a$ are uniform, running BP to convergence, and then modifying the factors to match those in the actual graph. From this viewpoint, the argument in Section 3 shows that an upper bound on the residual from uniform messages is

$$r(m_{cd}) \leq \max_{x_c} \left|\log \frac{t_c(x_c)}{u_c(x_c)}\right|, \quad (13)$$

where $u_c$ is a normalized uniform factor over the variables in $x_c$. Therefore, we use this upper bound as the initial priority of each update.

Finally, we need a way to approximate the residuals if damping is used. The important point here is that when a message $m_{bc}$ is sent with damping, even after the update is performed, the residual $m_{bc}$ is nonzero, because the full update has not been taken. This can be handled, however, by the following. Whenever a damped message $m_{cd}$ is sent, the residual $r(m_{bc})$ is computed exactly and $m_{bc}$ is added to the queue with that priority. (For simplicity, this is not shown in Algorithm 2.)

### 4.3 APPLICATION TO NON-INFERENCE DOMAINS

RBP1L has the advantage of being more general: it can readily be applied to any set of fixed-point equations, potentially ones that are very different than those used in approximate inference. On the other hand, RBP0L appears to be more specific to BP, because the residual bounds assume that BP updates are being used. For similar algorithms, such as max-product BP and GBP, it is likely that the same scheme would be effective. For a completely different set of fixed-point equations, applying RBP0L would require both designing a new method for approximating the update residuals, and designing an efficient way for initializing the residual updates. That said, our residual estimation procedure, which simply sums up the antecedent residuals, is fairly generic, and thus likely to perform well in a variety of domains.

## 5　EXPERIMENTS

In this section, we compare the convergence speed of RBP0L and RBP1L on both synthetic and real-world graphs.

### 5.1　SYNTHETIC DATA

We randomly generate $N \times N$ grids of binary variables with pairwise Potts factors. Each pairwise factor has the form

$$t_{ij}(x_i, x_j) = \begin{pmatrix} 1 & e^{-\alpha_{ij}} \\ e^{-\alpha_{ij}} & 1 \end{pmatrix}, \quad (14)$$

where the equality strength $\alpha$ is sampled uniformly from $[-C, C]$. Higher values of $C$ make inference more difficult. The unary factors have the form $t_i(x_i) = \begin{bmatrix} 1 & e^{-u_i} \end{bmatrix}$, where $u_i$ is sampled uniformly from $[-C, C]$. We generate 50 distributions for $C = 5$. For smaller values of $C$, inference becomes so easy that all schedules performed equally well. For larger values of $C$, the same trend holds, but the the convergence rates are much lower. We use the grid size $N = 10$ so that exact inference is still feasible. We measure running time by the number of message updates computed. This measure closely matches the CPU time. Both algorithms are considered to have converged when no pending update has a residual of greater than $10^{-3}$. The algorithms are considered to have diverged if they have not converged after the equivalent of 1000 complete sweeps of the graph.

The rate of convergence of the different schedules are shown in Figure 1. We see that RBP0L converges much more rapidly than RBP1L, although both eventually converge on the same percentage of networks.

Figure 2 shows the number of messages required for convergence for each sampled model. Each integer on the x-axis represents a different randomly-generated model, sorted by the number of messages required by RBP1L. Thus, the model at x-index 0 is the easiest model for RBP1L, and so on. Each curve is the number of messages required, as a function of this rank. The horizontal line is the number-of-messages cutoff, so points that exceed that line represent models for which BP did not converge. The y-axis is logarithmic.

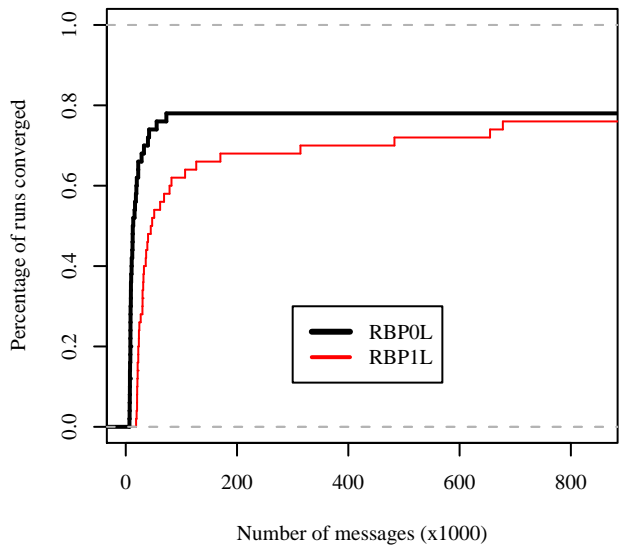

Figure 1: Convergence of RBP0L and RBP1L on synthetic $10 \times 10$ grids with $C = 5$. The x-axis is number of messages computed. RBP0L converges faster.

RBP0L computes on average half as many messages as RBP1L. RBP0L uses fewer messages than RBP1L in 46 of the 50 sampled models. In three of the sampled models, RBP1L converges but RBP0L does not, which appear in Figure 2 as the peaks where the RBP0L curve is the only one that touches the horizontal line. In three other models, RBP0L converges but RBP1L does not, which appear as the valleys where RBP0L does not touch the horizontal line, but the other curves do. The dashed curve in the figure shows the number of updates actually performed by RBP1L. On average, 38% of the updates computed by RBP1L are never performed. Surprisingly, RBP0L performs fewer updates than RBP1L performs; that is, it is more efficient even if wasted updates are not counted against RBP1L. This may be a beneficial effect of our choice of initial residual estimates.

Finally, we measure the accuracy of the marginals for RBP0L and RBP1L. For both schedules, we measure the average per-variable KL from the exact distribution to the BP belief. When both schedules converge, the average per-variable KL is nearly identical: the mean absolute difference, averaged over the 50 random models, is 0.0038.

### 5.2 NATURAL-LANGUAGE DATA

Finally, we consider a model with many irregular loops, which is the skip chain conditional random field introduced by Sutton and McCallum [2004]. This model incorporates certain long-distance dependencies between word labels into a linear-chain model for information extraction. The resulting networks contain many loops of varying sizes, and exact inference using a generic junction-tree solver is intractable.

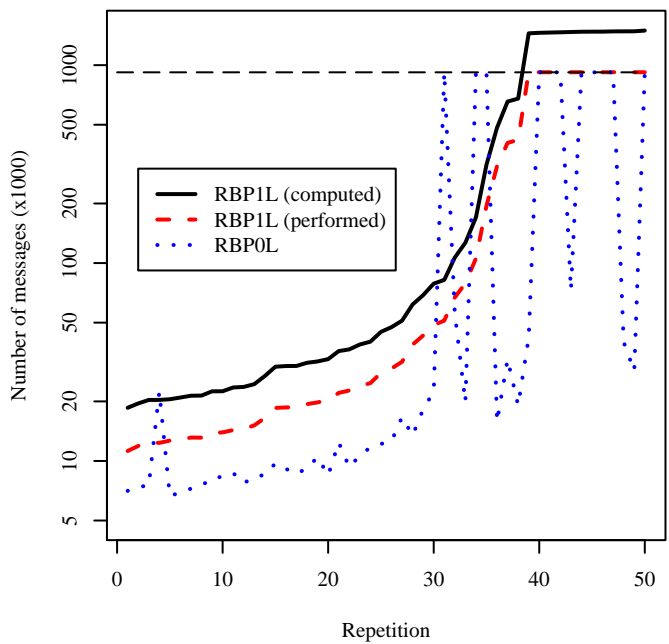

Figure 2: Updates performed by RBP0L and RBP1L on synthetic data. The horizontal line is the number-of-messages cutoff. The y-axis is logarithmic.

|        | Messages sent | Accuracy |
|--------|---------------|----------|
| TRP    | 3 079 570     | **97.6** |
| RBP0L  | **839 250**   | 97.4     |
| RBP1L  | 2 685 702     | 97.3     |

Table 1: Performance of BP schedules on skip-chain test data.

The task is to extract information about seminars from email announcements. Our data set is a collection of 485 e-mail messages announcing seminars at Carnegie Mellon University. The messages are annotated with the seminar's starting time, ending time, location, and speaker. This data set is due to Freitag [1998], and has been used in much previous work.

Often the speaker is listed multiple times in the same message. For example, the speaker's name might be included both near the beginning and later on, in a sentence such as "If you would like to meet with Professor Smith..." It can be useful to find both such mentions, because different information can be in the surrounding context of each mention: for example, the first mention might be near an institution affiliation, while the second mentions that Smith is a professor.

To increase recall of person names, we wish to exploit the fact that when the same word appears multiple times in the same message, it tends to have the same label. In a CRF, we can represent this by adding edges between output nodes $(y_i, y_j)$ when the words $x_i$ and $x_j$ are identical and capitalized.

The emails on average contain 273.1 tokens, but the maximum is 3062 tokens. The messages have an average of 23.5 skip edges, but the maximum is 2260, indicating that some networks are connected densely.

We generate networks as follows. Using ten-fold cross-validation with a 50/50 train/test split, we train a skip-chain CRF using TRP until the model parameters converge. Then we evaluate the RBP0L, RBP1L, and TRP on the test data, measuring the number of messages sent, the running time, and the accuracy on the test data. As in the last section, RBP0L and RBP1L are considered to have converged if no pending update has a residual of more than $10^{-3}$. TRP is considered to have converged if no update performed on the previous iteration resulted in a residual of greater than $10^{-3}$. In all cases, the trained model parameters are exactly the same; the inference algorithms are varied only at test time, not at training time.

Table 1 shows the performance of each of the message schedules, averaged over the 10 folds. RBP0L uses one-third of the messages as RBP1L, and one-fifth of the CPU time, but has essentially the same accuracy. Also, RBL0L uses 27% of the messages used by TRP.

In our implementation, the CPU time required per message update is much higher for the RBP schedules than for TRP. The total running time for RBP0L is 66s, compared to 110s for TRP and 321s for RBP1L. This is partially because of the overhead in maintaining the priority queues and residual estimates, but also this is because our TRP implementation is a highly optimized one that we have used in much previous work, whereas our RBP implementations have more room for low-level optimization.

### 5.3 ERROR ESTIMATES

The message residual is an intuitive error measure to use for scheduling, but there are many others that are conceivable. In this section, we compare different error measures to evaluate how reliable they are at predicting the next message to send. Ideally, we would evaluate a priority function for messages by whether higher priority messages actually reduces the computation time required for convergence. But it is extremely difficult to compute this, so we instead measure the distance to the converged message values, as follows.

We generate a synthetic grid as in Section 5.1. (The graphs here are from a single sampled model, but different samples result in qualitatively similar results.) Then, we run RBP0L on the grid to convergence, yielding a set of converged messages $\tilde{\mathbf{m}}$. Finally, we run RBP0L again on the same grid, without making use of $\tilde{\mathbf{m}}$. After each message update of RBP0L $m_{cd}^{(k)} \mapsto m_{cd}$, we measure: (a) The residual of the errors $e(m_{cd}^{(k)}, m_{cd})$, $e(m_{cd}^{(k)}, \tilde{m}_{cd})$, and $e(m_{cd}, \tilde{m}_{cd})$ (b) The dynamic range of the same errors (c) The KL divergences $KL(m_{cd}^{(k)} \| m_{cd})$, $KL(m_{cd}^{(k)} \| \tilde{m}_{cd})$, and $KL(m_{cd} \| \tilde{m}_{cd})$; (d) The change in Bethe energy $\log Z_{\mathrm{BP}}(\mathbf{m}) - \log Z_{\mathrm{BP}}(\mathbf{m}^{(k)})$.

Thus we can measure how well each of the error metrics predicts the distance to convergence $r(e(m_{cd}, \tilde{m}_c d) - r(e(m_{cd}^{(k)}, \tilde{m}_c d)$. This is shown in Figure 3. Each plot in that figure shows a different distance measure between messages: from top left, they are message residual, error dynamic range, KL divergence, and difference in Bethe energy. Each point in the figures represents a single message update. In all figures, the $x$-axis shows the distance between the message $m_{cd}^{(k)}$ at the previous iteration and the value $m^{()}(k+1)$ at the current iteration. The $y$-axis shows the change in distance to the converged messages, that is, how much closer the update at $k+1$ brought the message to its converged value. We measure this as the difference between the residuals $e(m_{cd}^{(k+1)}, \tilde{m}_{cd})$ and $e(m_{cd}^{(k)}, \tilde{m}_{cd})$. Negative values of this measure are better, because they mean that the distance to the converged messages has decreased due to the update. An ideal graph would be a line with negative slope.

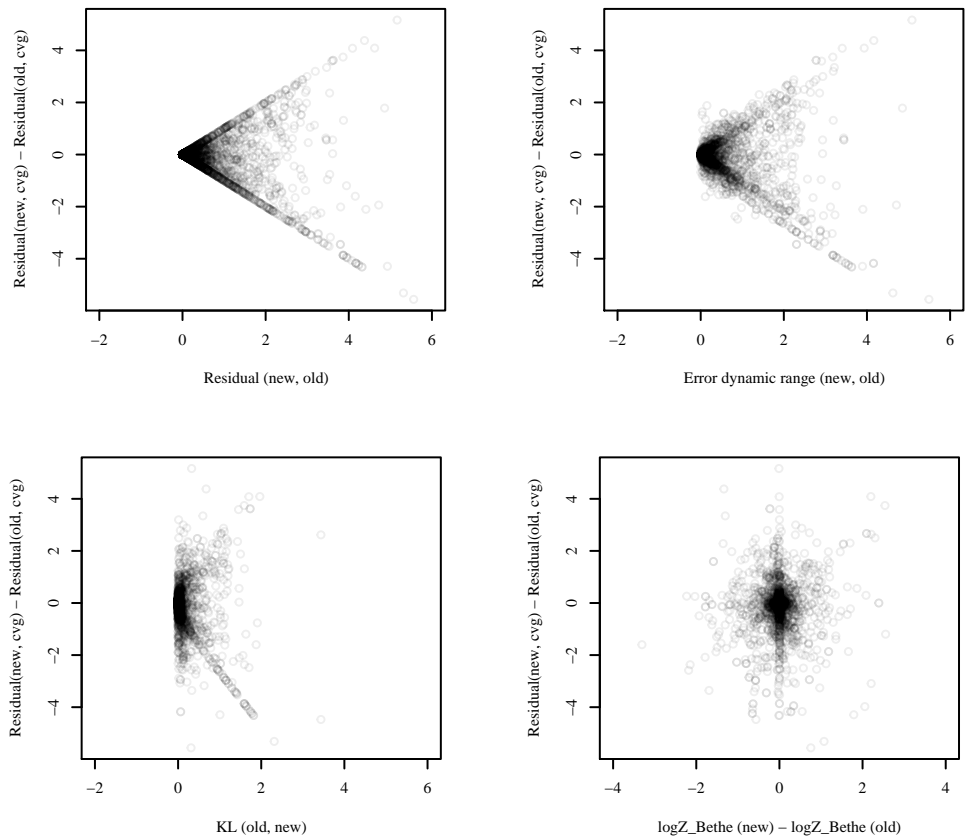

Figure 3: Comparison of error metrics in predicting the distance to convergence. (See text for explanation.)

Both the message residual and the dynamic error range display a clear upper-bounding property on the absolute value. Also, the points are somewhat clustered along the diagonal, indicating some kind of a linear relationship between the single-message distance and the distance to convergence. The single-message distance does not seem to do well, however, at predicting *in which direction* the message will change, that is, closer or farther from its converged value. Qualitatively, the residual and the error range seem to perform similarly at predicting the distance to the converged messages, but in preliminary experiments, using the error range in a one-lookahead schedule seemed to converge slightly slower than using the residual.

The message KL also seems to do a poor job of predicting the distance to the converged message. More surprisingly, the difference in Bethe energy is almost completely uninformative about the distance to converged messages. This suggests an intriguing explanation of the slow converge of gradient methods for optimizing the Bethe energy: perhaps the objective function itself is simply not good at measuring what we care about. It is possible that the Bethe approximation may be accurate at convergence but still not be accurate outside of the constraint set, that is, when the messages are not locally consistent. This is precisely the situation that occurs during message scheduling. For this reason, it may be more revealing to look at the Lagrangian of the Bethe energy rather than the objective function itself.

## 6 CONCLUSION

In this paper, we have presented RBP0L, a new dynamic schedule for belief propagation that schedules messages based on a upper-bound on their residual. On both synthetic and real-world data, we show that it converges faster than both RBP1L, a recently-proposed dynamic schedule, and than TRP, with comparable accuracy. It would be interesting to explore whether the residual estimation technique in RBP0L is equally effective for other inference algorithms, such as EP, GBP, or whether the residual estimation technique would require significant adaptation. In continuous spaces, it may be that the message residual itself is not a good measure for scheduling, because it gives equal weight to all areas of the domain, even those with low probability. The KL divergence may be more appropriate.

## A Appendix

In this appendix, we prove the upper bound (8) given in Section 3. This is

$$r(m_{cd}) \leq \sum_{\{b \in N(c)\}} r(m_{bc}) \tag{15}$$

To justify this, we show that the residual is both subadditive and contracts under the message update, following Ihler et al. [2004].

To show subadditivity, define the message product $M_{bc}(x_c) = \prod_{\{b \in N(c)\} \setminus d} m_{bc}(x_c)$, and define $M_{bc}^{(k)}$ similarly. Also, define the residual $r(M_{bc})$ as

$$r(M_{bc}) = \max_{x_c} \left| \log \frac{M_{bc}(x_c)}{M_{bc}^{(k)}(x_c)} \right| \tag{16}$$

Then we have

$$r(M_{bc}) = \max_{x_c} \left| \sum_b \log \frac{m_{bc}(x_c)}{m_{bc}^{(k)}(x_c)} \right|$$

$$\leq \sum_b \max_{x_c} \left| \log \frac{m_{bc}(x_c)}{m_{bc}^{(k)}(x_c)} \right| = \sum_b r(m_{bc}),$$

which follows from the subadditivity of absolute value, and an increase in the degrees of freedom of the maximization.

To show contraction under the message update, we apply the fact that

$$\frac{f_1 + f_2}{g_1 + g_2} \leq \max\left\{\frac{f_1}{g_1}, \frac{f_2}{g_2}\right\}. \tag{17}$$

This directly yields

$$r(m_{cd}) = \max_{x_{cd}} \left| \log \frac{\sum_{x_c \setminus x_{cd}} t_c(x_c) M_{bc}}{\sum_{x_c \setminus x_{cd}} t_c(x_c) M_{bc}^{(k)}} \right| \tag{18}$$

$$\leq \max_{x_{cd}} \left| \log \max_{x_c \setminus x_{cd}} \frac{t_c(x_c) M_{bc}}{t_c(x_c) M_{bc}^{(k)}} \right| \tag{19}$$

$$\leq \max_{x_{cd}} \max_{x_c \setminus x_{cd}} \left| \log \frac{M_{bc}(x_c)}{M_{bc}^{(k)}(x_c)} \right| \tag{20}$$

$$= r(M_{bc}). \tag{21}$$

## Acknowledgements

We thank the anonymous reviewers for detailed and helpful comments. This work was supported in part by the Center for Intelligent Information Retrieval and in part by The Central Intelligence Agency, the National Security Agency and National Science Foundation under NSF grants #IIS-0326249 and #IIS-0427594. Any opinions, findings and conclusions or recommendations expressed in this material are the authors' and do not necessarily reflect those of the sponsor.